\documentclass[11pt,a4paper]{article}
\usepackage[hyperref]{acl2019}
\usepackage{times}
\usepackage{latexsym}
\usepackage{url}
\usepackage{subfigure}

\aclfinalcopy 

\usepackage{amsmath}
\usepackage{amssymb}
\usepackage{booktabs}
\usepackage{multirow}
\usepackage{lipsum}
\usepackage{xfrac}

\usepackage{ulem} 
\usepackage{tikz}
\usetikzlibrary{arrows,calc,decorations.pathreplacing,positioning}

\usepackage{CJKutf8}

\usepackage{enumitem}
\setlist{nosep}

\DeclareMathOperator*{\argmax}{argmax}

\title{An Empirical Study of Generation Order for Machine Translation}

\author{William Chan\Thanks{\;Equal contribution.} \\
  Google Research, Brain Team\\
  \texttt{williamchan@google.com}\\\And
  Mitchell Stern\footnotemark[1] \\
  University of California, Berkeley \\
  \texttt{mitchell@berkeley.edu} \\\AND
  Jamie Kiros \\
  Google Research, Brain Team \\
  \texttt{kiros@google.com} \\\And
  Jakob Uszkoreit \\
  Google Research, Brain Team\\
  \texttt{uszkoreit@google.com} \\}

\date{}

\begin{document}

\maketitle

\begin{abstract}
In this work, we present an empirical study of generation order for machine translation. Building on recent advances in insertion-based modeling, we first introduce a soft order-reward framework that enables us to train models to follow arbitrary oracle generation policies. We then make use of this framework to explore a large variety of generation orders, including uninformed orders, location-based orders, frequency-based orders, content-based orders, and model-based orders. Curiously, we find that for the WMT'14 English $\to$ German translation task, order does not have a substantial impact on output quality, with unintuitive orderings such as alphabetical and shortest-first matching the performance of a standard Transformer. This demonstrates that traditional left-to-right generation is not strictly necessary to achieve high performance. On the other hand, results on the WMT'18 English $\to$ Chinese task tend to vary more widely, suggesting that translation for less well-aligned language pairs may be more sensitive to generation order.

\end{abstract}

\section{Introduction}

Neural sequence models \cite{sutskever-nips-2014,cho-emnlp-2014} have been successfully applied to a broad range of tasks in recent years.
While these models typically generate their outputs using a fixed left-to-right order, there has also been some investigation into non-left-to-right and order-independent generation in pursuit of quality or speed. For example, \citet{vinyals-iclr-2015} explored the problem of predicting sets using sequence models. While this is a domain where generation order should intuitively be unimportant, they nevertheless found it to make a substantial difference in practice. \citet{ford-emnlp-2018} explored treating language modeling as a two-pass process, where words from certain classes are generated first, and the remaining words are filled in during the second pass. They found that generating function words first followed by content words second yielded the best results. Separately, \citet{gu-iclr-2018} and \citet{lee-emnlp-2018} developed non-autoregressive approaches to machine translation where the entire output can be generated in parallel in constant time. These models do away with order selection altogether but typically lag behind their autoregressive counterparts in translation quality.

More recently, a number of novel insertion-based architectures have been developed for sequence generation \cite{gu-arxiv-2019,stern-arxiv-2019,welleck-arxiv-2019}. These frameworks license a diverse set of generation orders, including uniform \cite{welleck-arxiv-2019}, random \cite{gu-arxiv-2019}, or balanced binary trees \cite{stern-arxiv-2019}. Some of them also match the quality of state-of-the-art left-to-right models \cite{stern-arxiv-2019}.
In this paper, we utilize one such framework to explore an extensive collection of generation orders, evaluating them on the WMT'14 English-German and WMT'18 English-Chinese translation tasks. We find that a number of non-standard choices achieve BLEU scores comparable to those obtained with the classical approach, suggesting that left-to-right generation might not be a necessary ingredient for high-quality translation. Our contributions are as follows:



\begin{figure*}[t]
\centering
\resizebox{0.75\width}{!}{%
\begin{CJK*}{UTF8}{gbsn}
\begin{tikzpicture}[xscale=1.4, yscale=1.3]
\tikzstyle{layer} = [draw=black, rectangle, rounded corners, thick, minimum width=7cm, minimum height=0.75cm, align=center];
\tikzstyle{arrow} = [->, >=triangle 45];
\draw[decorate, decoration={brace,amplitude=5pt,mirror,raise=4ex}] (-0.5, 0) -- (4.5, 0) node[midway,yshift=-3em]{Encoder};
\node (zh-word-1) at (0, 0) {那个};
\node (zh-word-2) at (1, 0) {男人};
\node (zh-word-3) at (2, 0) {吃};
\node (zh-word-4) at (3, 0) {了};
\node (zh-word-5) at (4, 0) {小吃};
\node[layer] (zh-layer-1) at (2, 1) {Embedding};
\node[layer] (zh-layer-2) at (2, 2) {Self-Attention};
\draw[arrow] (zh-word-1.north) -- (zh-word-1.north |- zh-layer-1.south);
\draw[arrow] (zh-word-2.north) -- (zh-word-2.north |- zh-layer-1.south);
\draw[arrow] (zh-word-3.north) -- (zh-word-3.north |- zh-layer-1.south);
\draw[arrow] (zh-word-4.north) -- (zh-word-4.north |- zh-layer-1.south);
\draw[arrow] (zh-word-5.north) -- (zh-word-5.north |- zh-layer-1.south);
\draw[arrow] (zh-layer-1.north) -- (zh-layer-2.south);
\begin{scope}[shift={(6, 0)}]
\draw[decorate, decoration={brace,amplitude=5pt,mirror,raise=4ex}] (-0.5, 0) -- (4.5, 0) node[midway,yshift=-3em]{Decoder};
\node (en-word-1) at (0, 0) {$\langle$START$\rangle$};
\node (en-word-2) at (1.33, 0) {ate};
\node (en-word-3) at (2.66, 0) {snack};
\node (en-word-4) at (4, 0) {$\langle$END$\rangle$};
\node (en-target-1) at (0.75, 4) {\{the, man\}};
\node (en-target-2) at (2, 4) {\{a\}};
\node (en-target-3) at (3.25, 4) {\{\}};
\node[layer] (en-layer-1) at (2, 1) {Embedding};
\node[layer] (en-layer-2) at (2, 2) {Self-Attention + Cross-Attention};
\node[layer] (en-layer-3) at (2, 3) {Content and Location Softmaxes};
\draw[arrow] (en-word-1.north) -- (en-word-1.north |- en-layer-1.south);
\draw[arrow] (en-word-2.north) -- (en-word-2.north |- en-layer-1.south);
\draw[arrow] (en-word-3.north) -- (en-word-3.north |- en-layer-1.south);
\draw[arrow] (en-word-4.north) -- (en-word-4.north |- en-layer-1.south);
\draw[arrow] (en-layer-1.north) -- (en-layer-2.south);
\draw[arrow] (en-layer-2.north) -- (en-layer-3.south);
\draw[arrow] (en-layer-3.north -| en-target-1) -- (en-target-1.south);
\draw[arrow] (en-layer-3.north -| en-target-2) -- (en-target-2.south);
\draw[arrow] (en-layer-3.north -| en-target-3) -- (en-target-3.south);
\end{scope}
\draw (zh-layer-2.north) |- (5, 2.5) |- ($(en-layer-1.north)!0.5!(en-layer-2.south)$);
\end{tikzpicture}
\end{CJK*}
}
\caption{A schematic of the Insertion Transformer model for a Chinese-English translation pair. The model is encouraged to predict the correct set of remaining words within each slot. Using our order-reward framework (Section~\ref{sec:reward-framework}), we can derive the necessary weight distribution to apply to the set of correct actions in order to train the model to follow any oracle generation policy of interest.}
\label{fig:insertion-transformer-model}
\end{figure*}
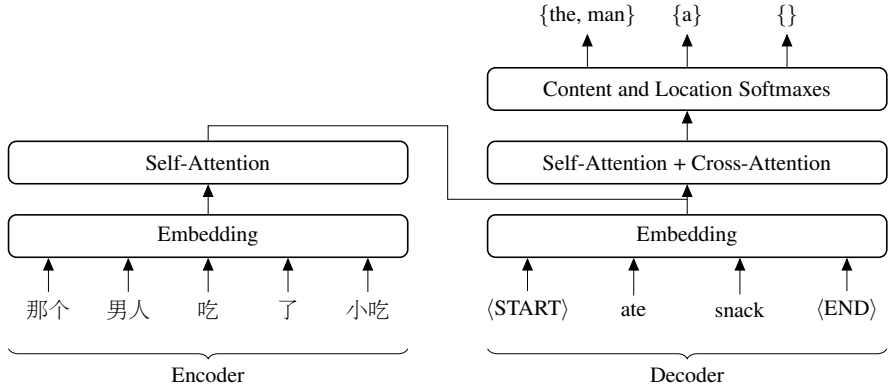

\begin{figure*}[t]
\hfill
\begin{minipage}[t]{0.44\textwidth}
\centering
\vspace{0pt}
Serial generation: \\
\vspace{0.5em}
\begin{tabular}{ll}
\toprule
Hypothesis & Insertion \\
\midrule
{ [] } & (ate, 0) \\
{ [ate] } & (snack, 1) \\
{ [ate, snack] } & (man, 0) \\
{ [man, ate, snack] } & (the, 0) \\
{ [the, man, ate, snack] } & (a, 3) \\
{ [the, man, ate, a, snack] } & ($\langle$EOS$\rangle$, 5) \\
\bottomrule
\end{tabular}
\end{minipage}
\hfill
\begin{minipage}[t]{0.53\textwidth}
\centering
\vspace{0pt}
Parallel generation: \\
\vspace{0.5em}
\begin{tabular}{ll}
\toprule
Hypothesis & Insertions \\
\midrule
{ [] } & (ate, 0) \\
{ [ate] } & (man, 0), (snack, 1) \\
{ [man, ate, snack] } & (the, 0), (a, 2) \\
{ [the, man, ate, a, snack] } & ($\langle$EOS$\rangle$, 5) \\
\bottomrule
\end{tabular}
\end{minipage}
\hfill
\caption{Example decoding paths for serial and parallel generation using the Insertion Transformer.}
\label{fig:insertion-transformer-generation}
\end{figure*}

\begin{itemize}[leftmargin=*,noitemsep]
\item We introduce a general soft order-reward framework that can be used to teach insertion-based models to follow any specified ordering.
\item We perform a thorough empirical study of various orders, including: uniform, random, left-to-right, right-to-left, common-first, rare-first, shortest-first, longest-first, alphabetical, and model-adaptive.
\item On the WMT 2014 English $\rightarrow$ German task, we show that there is surprisingly little variation in BLEU for different generation orders. We further find that many orders are able to match the performance of a standard base Transformer.
\end{itemize}

\section{Background}

Neural sequence models have traditionally been designed with left-to-right prediction in mind. In the classical setting, output sequences are produced by repeatedly appending tokens to the rightmost end of the hypothesis until an end-of-sequence token is generated. Though high-performing across a wide range of application areas, this approach lacks the flexibility to accommodate other types of inference such as parallel generation, constrained decoding, infilling, etc. Moreover, it also leaves open the possibility that a non-left-to-right factorization of the joint distribution over output sequences could outperform the usual monotonic ordering.

To address these concerns, several recent approaches have been proposed for insertion-based sequence modeling, in which sequences are constructed by repeatedly inserting tokens at arbitrary locations in the output rather than only at the right-most position. We use one such insertion-based model, the Insertion Transformer \cite{stern-arxiv-2019}, for our empirical study. We give a brief overview of the model in this section before moving on to the details of our investigation.

\subsection{Insertion Transformer}

The Insertion Transformer \cite{stern-arxiv-2019} is a sequence-to-sequence model in which the output is formed by successively inserting one or more tokens at arbitrary locations into a partial hypothesis.
This type of generation is made possible through the use of a joint distribution over tokens and slots.
More formally, given an input $x$ and a partial output $\hat{y}_t$ at time $t$, the Insertion Transformer gives the joint distribution
\begin{align*}
p(c, l \mid x, \hat{y}_t) = \mathrm{InsertionTransformer}(x, \hat{y}_t) ,
\end{align*}
where $c \in V$ is the content being selected from the vocabulary $V$ and $0 \le l \le |\hat{y}_t|$ is the insertion location.

As its name suggests, the Insertion Transformer extends the Transformer model \cite{vaswani-nips-2017} with a few key modifications to generalize from ordinary next-token modeling to joint token-and-slot modeling. First, the Insertion Transformer removes the causal attention mask from the decoder, allowing for fully contextualized output representations to be derived after each insertion. Second, the Insertion Transformer pads the length-$n$ decoder input on both ends so that $n + 2$ output vectors are produced. It then concatenates adjacent pairs of output vectors to obtain $n + 1$ slot representations, which in turn inform the conditional distributions over tokens within each slot, $p(c \mid l)$. Lastly, it performs an additional attention step over the slot representations to obtain a location distribution $p(l)$, which is multiplied with the conditional content distributions to obtain the full joint distribution: $p(c, l) = p(c \mid l) p(l)$. A schematic of the architecture is given in Figure~\ref{fig:insertion-transformer-model} for reference.

We note that \citet{stern-arxiv-2019} also experimented with a number of other architectural variants, but we use the baseline version of the model described above in our experiments for simplicity.

\subsection{Decoding}

Once the model has been trained, it can be used for greedy autoregressive sequence generation as follows. At each step of decoding, we compute the joint $\argmax$ $$(\hat{c}_t, \hat{l}_t) = \argmax_{c, l} p(c, l \mid x, \hat{y}_t)$$ to determine what content $\hat{c}_t$ should be inserted at which location $\hat{l}_t$. We then apply this insertion, increasing the sequence length by one, and repeat this process until an end-of-sequence token is produced. This is the serial decoding procedure shown in the left half of Figure~\ref{fig:insertion-transformer-generation}.

The model can also be used for parallel partially-autoregressive decoding. Instead of computing the joint $\argmax$ across all locations, we instead compute the best content for each location: $$\hat{c}_{l, t} = \argmax_c p(c \mid l, x, \hat{y}_t) .$$ We then insert the highest-scoring tokens in parallel for all slots that are not yet finished, increasing the sequence length by anywhere between one and $n + 1$ tokens. This strategy visualized in the right half of Figure~\ref{fig:insertion-transformer-generation}.

\section{Soft Order-Reward Framework}
\label{sec:reward-framework}

\begin{table}[t]
\centering
\resizebox{\columnwidth}{!}{%
\begin{tabular}{ll}
\toprule
\bfseries Order & \bfseries Order Function $O(a)$ \\
\midrule
Uniform & $0$ \\
Balanced Binary Tree & $\left| s - (i + j) / 2 \right|$ \\
Random & $\mathrm{rank}(\mathrm{hash}(w))$ \\
Sequential (L2R vs.\ R2L) & $\pm s$ \\
Frequency (Common vs.\ Rare) & $\pm \mathrm{rank}(\mathrm{frequency}(w))$ \\
Length (Short vs.\ Long) & $\pm \mathrm{rank}(\mathrm{length}(w))$ \\
Alphabetical (A $\rightarrow$ z vs.\ z $\rightarrow$ A) & $\pm \mathrm{rank}(w)$ \\
Adaptive (Easy vs.\ Hard) & $\pm \log p(a)$ \\
\bottomrule
\end{tabular}
}
\caption{Order functions for an action $a$ corresponding to the insertion of word $w$ into slot $s$ within span $(i, j)$. The $\mathrm{rank}$ terms are computed with respect to the set of words from the valid action set $A^*$.}
\label{table:order}
\end{table}

Having presented our model of interest, we now describe a general soft order-reward framework that can be used to train the model to follow any oracle ordering for sequence generation. Let $O(a)$ be an order function mapping insertion actions to real numbers, where lower values correspond to better actions, and let $p(a)$ be the probability assigned by the model to action $a$. From these, we construct a reward function $R(a)$, an oracle policy $q_{\mathrm{oracle}}$, and a per-slot loss $\mathcal{L}$:
\begin{align*}
    R(a) &= \begin{cases}
        -O(a) & \forall a \in A^* \\
        -\infty & \forall a \not\in A^*
    \end{cases} \\
    q_{\mathrm{oracle}}(a) &= \frac{\exp(R(a) / \tau)}{\sum_{a' \in A^*} \exp(R(a') / \tau)} \\ 
    \mathcal{L} &= \mathrm{KL}(q_{\mathrm{oracle}} \, \| \, p) 
\end{align*}
Here, $A^*$ is the set of all valid actions. The temperature $\tau \in (0, \infty)$ controls the sharpness of the distribution, where $\tau \to 0$ results in a one-hot distribution with all mass on the best-scoring action under the order function $O(a)$, and $\tau \rightarrow \infty$ results in a uniform distribution over all valid actions. Intermediate values of $\tau$ result in distributions which are biased towards better-scoring actions but allow for other valid actions to be taken some of the time.

Having defined the target distribution, we take the slot loss $\mathcal{L}$ for insertions within a particular slot to be the KL-divergence between the oracle distribution $q_{\mathrm{oracle}}$ and the model distribution $p$. Substituting $\mathcal{L}$ in for the slot loss within the training framework of \citet{stern-arxiv-2019} then gives the full sequence generation loss, which we can use to train an Insertion Transformer under any oracle policy rather than just the specific one they propose. We describe a wide variety of generation orders which can be characterized by different order functions $O(a)$ in the subsections that follow. A summary is given in Table~\ref{table:order}.

\subsection{Uninformed Orders}
We evaluate two uninformed orders, uniform and random. The uniform order $O(a) = 0$ gives equal reward or equivalently probability mass to any valid action. Consequently, this means we give each order a uniform probability treatment. We also experiment with random order $O(a) = \mathrm{rank}(\mathrm{hash}(w))$, wherein we hash each word and use the sorted hash ID as the generation order. The random order forces the model to follow a specific random path, whereas the uniform order gives equal probability mass to any order.

\subsection{Location-based Orders}
We explore two types of location-based orders, balanced binary tree and monotonic orders. The balanced binary tree order $O(a) = |s - (i + j) / 2|$ encourages the model to place most of its probability mass towards the middle tokens in a missing span. Consequently, this encourages the model to generate text in a balanced binary tree order. We also experiment with soft monotonic orders $O(a) = \pm s$, or soft left-to-right and soft right-to-left, which differ slightly from the left-to-right teacher forcing traditionally used in seq2seq. First, we still maintain a uniform roll-in policy (see Section~\ref{sec:rollin}), which increases diversity during training and helps avoid label bias. Additionally, this endows the model with the ability to ``look back'' and insert missing tokens in the middle of the sequence during inference, as opposed to always being forced to append only at one end of the sequence. The order reward is also soft (as described by the $\tau$ term above), wherein we do not place all the probability mass on the next monotonic token, but merely encourage it to generate in a monotonic fashion.

\subsection{Frequency-based Orders}
We evaluate two frequency-based orders: rare words first via $O(a) = \mathrm{rank}(\mathrm{frequency}(w))$ and common words first via $O(a) = -\mathrm{rank}(\mathrm{frequency}(w))$. For these orders, we simply sort the words based on their frequencies and used their rank as the order. We note the most frequent words tend to be punctuation and stop words, such as commas, periods, and ``the'' in English.

\subsection{Content-based Orders}
We also explore content-based orders. One class of orders is based on the word length: $O(a) = \pm \mathrm{rank}(\mathrm{length}(w))$. This encourages the model to either emit all the shortest words first or all the longest words first.

We also explore alphabetical orderings $O(a) = \pm \mathrm{rank}(w)$, where sorting is based on Unicode order.
We note that in Unicode, uppercase letters occur before lower case letters. This biases the model to produce words which are capitalized first (or last), typically corresponding to nouns in German. Additionally, for Chinese, the characters are roughly sorted by radical and stroke count, which bears a loose relation to the complexity and frequency of the character.

\subsection{Model-based Orders}
The orders presented thus far are static, meaning they are independent of the model. We also explore orders which are adaptive based on the model's posterior.
We also introduce ``easy'' and ``hard'' adaptive orders induced by $O(a) = \pm \log p(a)$. The adaptive orders look at the model's posterior to determine the oracle policy. Consequently the loss is adaptive, as when the model updates after each gradient step, the order adapts to the model's posterior.

In the ``easy`` version, we use $O(a) = +\log p(a)$, which is similar to a local greedy soft EM loss. We renormalize our current model's posterior over valid actions and optimize towards that distribution. This pushes the model's posterior to what is correct and where it has already placed probability mass.
Intuitively, this reinforces the model to select what it thinks are the easiest actions first.
Conversely, the ``hard'' variant uses $O(a) = -\log p(a)$ which encourages the model to place probability mass on what it thinks are the hardest valid actions. This is akin to a negative feedback system whose stationary condition is the uniform distribution.

\subsection{Roll-in Policy}
\label{sec:rollin}

We follow \citet{stern-arxiv-2019} and use a uniform roll-in policy when sampling partial outputs at training time in which we first select a subset size uniformly at random, then select a random subset of the output of that size.
Repeated tokens are handled via greedy left or right alignment to the true output.

\begin{figure*}[t]
\small
\begin{flushleft}
\textbf{Input:} It would of course be a little simpler for the Germans if there were a coherent and standardised European policy, which is currently not the case. \\
\vspace{0.5em}
\textbf{Output:} Es wäre für die Deutschen natürlich ein wenig einfacher, wenn es eine kohärente und einheitliche europäische Politik gäbe, was derzeit nicht der Fall ist. \\
\vspace{0.5em}
\textbf{Parallel decode (alphabetical):}
\end{flushleft}
\vspace{-1.5em}
\begin{center}
\resizebox{\textwidth}{!}{%
\begin{tikzpicture}[every node/.style={anchor=base,inner sep=0.1em}]
\input{example-german-alphabetical.tikz}
\end{tikzpicture}
}
\end{center}
\vspace{-0.5em}
\begin{CJK*}{UTF8}{gbsn}
\begin{flushleft}
\textbf{Input:} according to the data of National Bureau of Statistics , the fixed asset investment growth , total imports and other data in July have come down . \\
\vspace{0.5em}
\textbf{Output:} 根据 国家统计局 的 数据 ， 7 月份 的 固定资产 投资 增长 、 进口 总额 和 其他 数据 有所 下降 。 \\
\vspace{0.5em}
\textbf{Parallel decode (alphabetical):}
\end{flushleft}
\vspace{-0.5em}
\begin{flushleft}
\scriptsize
\begin{tikzpicture}[every node/.style={anchor=base,inner sep=0.1em}]
\input{example-chinese-alphabetical.tikz}
\end{tikzpicture}
\end{flushleft}
\end{CJK*}
\caption{Example decodes for models trained to generate tokens in alphabetical (Unicode) order. Blue tokens correspond those being inserted at the current time step, and gray tokens correspond to those not yet generated. Note that the desired ordering applies on a per-slot basis rather than a global basis.}
\label{fig:alphabetical-examples}
\end{figure*}

\begin{figure}[t]
\small
\begin{flushleft}
\textbf{Input:} It will be sung by all the artists at all the three concerts at the same time. \\
\vspace{0.5em}
\textbf{Output:} Es wird von allen Künstlern bei allen drei Konzerten gleichzeitig gesungen. \\
\vspace{0.5em}
\textbf{Parallel decode (longest-first):}
\end{flushleft}
\vspace{-1.5em}
\begin{flushleft}
\resizebox{\columnwidth}{!}{%
\begin{tikzpicture}[every node/.style={anchor=base,inner sep=0.1em}]
\node[gray!60] (node-0-0) {Es\_};
\node[gray!60,base right=0cm of node-0-0] (node-0-1) {wird\_};
\node[gray!60,base right=0cm of node-0-1] (node-0-2) {von\_};
\node[gray!60,base right=0cm of node-0-2] (node-0-3) {allen\_};
\node[gray!60,base right=0cm of node-0-3] (node-0-4) {Künstler};
\node[gray!60,base right=0cm of node-0-4] (node-0-5) {n\_};
\node[gray!60,base right=0cm of node-0-5] (node-0-6) {bei\_};
\node[gray!60,base right=0cm of node-0-6] (node-0-7) {allen\_};
\node[gray!60,base right=0cm of node-0-7] (node-0-8) {drei\_};
\node[gray!60,base right=0cm of node-0-8] (node-0-9) {Konzert};
\node[gray!60,base right=0cm of node-0-9] (node-0-10) {en\_};
\node[blue,base right=0cm of node-0-10] (node-0-11) {\uline{gleichzeitig\_}};
\node[gray!60,base right=0cm of node-0-11] (node-0-12) {ges};
\node[gray!60,base right=0cm of node-0-12] (node-0-13) {ungen\_};
\node[gray!60,base right=0cm of node-0-13] (node-0-14) {.\_};
\node[gray!60,below=0.1cm of node-0-0] (node-1-0) {Es\_};
\node[gray!60,base right=0cm of node-1-0] (node-1-1) {wird\_};
\node[gray!60,base right=0cm of node-1-1] (node-1-2) {von\_};
\node[gray!60,base right=0cm of node-1-2] (node-1-3) {allen\_};
\node[blue,base right=0cm of node-1-3] (node-1-4) {\uline{Künstler}};
\node[gray!60,base right=0cm of node-1-4] (node-1-5) {n\_};
\node[gray!60,base right=0cm of node-1-5] (node-1-6) {bei\_};
\node[gray!60,base right=0cm of node-1-6] (node-1-7) {allen\_};
\node[gray!60,base right=0cm of node-1-7] (node-1-8) {drei\_};
\node[gray!60,base right=0cm of node-1-8] (node-1-9) {Konzert};
\node[gray!60,base right=0cm of node-1-9] (node-1-10) {en\_};
\node[black,base right=0cm of node-1-10] (node-1-11) {gleichzeitig\_};
\node[gray!60,base right=0cm of node-1-11] (node-1-12) {ges};
\node[blue,base right=0cm of node-1-12] (node-1-13) {\uline{ungen\_}};
\node[gray!60,base right=0cm of node-1-13] (node-1-14) {.\_};
\node[gray!60,below=0.1cm of node-1-0] (node-2-0) {Es\_};
\node[gray!60,base right=0cm of node-2-0] (node-2-1) {wird\_};
\node[gray!60,base right=0cm of node-2-1] (node-2-2) {von\_};
\node[blue,base right=0cm of node-2-2] (node-2-3) {\uline{allen\_}};
\node[black,base right=0cm of node-2-3] (node-2-4) {Künstler};
\node[gray!60,base right=0cm of node-2-4] (node-2-5) {n\_};
\node[gray!60,base right=0cm of node-2-5] (node-2-6) {bei\_};
\node[gray!60,base right=0cm of node-2-6] (node-2-7) {allen\_};
\node[gray!60,base right=0cm of node-2-7] (node-2-8) {drei\_};
\node[blue,base right=0cm of node-2-8] (node-2-9) {\uline{Konzert}};
\node[gray!60,base right=0cm of node-2-9] (node-2-10) {en\_};
\node[black,base right=0cm of node-2-10] (node-2-11) {gleichzeitig\_};
\node[blue,base right=0cm of node-2-11] (node-2-12) {\uline{ges}};
\node[black,base right=0cm of node-2-12] (node-2-13) {ungen\_};
\node[blue,base right=0cm of node-2-13] (node-2-14) {\uline{.\_}};
\node[gray!60,below=0.1cm of node-2-0] (node-3-0) {Es\_};
\node[blue,base right=0cm of node-3-0] (node-3-1) {\uline{wird\_}};
\node[gray!60,base right=0cm of node-3-1] (node-3-2) {von\_};
\node[black,base right=0cm of node-3-2] (node-3-3) {allen\_};
\node[black,base right=0cm of node-3-3] (node-3-4) {Künstler};
\node[gray!60,base right=0cm of node-3-4] (node-3-5) {n\_};
\node[gray!60,base right=0cm of node-3-5] (node-3-6) {bei\_};
\node[blue,base right=0cm of node-3-6] (node-3-7) {\uline{allen\_}};
\node[gray!60,base right=0cm of node-3-7] (node-3-8) {drei\_};
\node[black,base right=0cm of node-3-8] (node-3-9) {Konzert};
\node[blue,base right=0cm of node-3-9] (node-3-10) {\uline{en\_}};
\node[black,base right=0cm of node-3-10] (node-3-11) {gleichzeitig\_};
\node[black,base right=0cm of node-3-11] (node-3-12) {ges};
\node[black,base right=0cm of node-3-12] (node-3-13) {ungen\_};
\node[black,base right=0cm of node-3-13] (node-3-14) {.\_};
\node[blue,below=0.1cm of node-3-0] (node-4-0) {\uline{Es\_}};
\node[black,base right=0cm of node-4-0] (node-4-1) {wird\_};
\node[blue,base right=0cm of node-4-1] (node-4-2) {\uline{von\_}};
\node[black,base right=0cm of node-4-2] (node-4-3) {allen\_};
\node[black,base right=0cm of node-4-3] (node-4-4) {Künstler};
\node[gray!60,base right=0cm of node-4-4] (node-4-5) {n\_};
\node[blue,base right=0cm of node-4-5] (node-4-6) {\uline{bei\_}};
\node[black,base right=0cm of node-4-6] (node-4-7) {allen\_};
\node[blue,base right=0cm of node-4-7] (node-4-8) {\uline{drei\_}};
\node[black,base right=0cm of node-4-8] (node-4-9) {Konzert};
\node[black,base right=0cm of node-4-9] (node-4-10) {en\_};
\node[black,base right=0cm of node-4-10] (node-4-11) {gleichzeitig\_};
\node[black,base right=0cm of node-4-11] (node-4-12) {ges};
\node[black,base right=0cm of node-4-12] (node-4-13) {ungen\_};
\node[black,base right=0cm of node-4-13] (node-4-14) {.\_};
\node[black,below=0.1cm of node-4-0] (node-5-0) {Es\_};
\node[black,base right=0cm of node-5-0] (node-5-1) {wird\_};
\node[black,base right=0cm of node-5-1] (node-5-2) {von\_};
\node[black,base right=0cm of node-5-2] (node-5-3) {allen\_};
\node[black,base right=0cm of node-5-3] (node-5-4) {Künstler};
\node[blue,base right=0cm of node-5-4] (node-5-5) {\uline{n\_}};
\node[black,base right=0cm of node-5-5] (node-5-6) {bei\_};
\node[black,base right=0cm of node-5-6] (node-5-7) {allen\_};
\node[black,base right=0cm of node-5-7] (node-5-8) {drei\_};
\node[black,base right=0cm of node-5-8] (node-5-9) {Konzert};
\node[black,base right=0cm of node-5-9] (node-5-10) {en\_};
\node[black,base right=0cm of node-5-10] (node-5-11) {gleichzeitig\_};
\node[black,base right=0cm of node-5-11] (node-5-12) {ges};
\node[black,base right=0cm of node-5-12] (node-5-13) {ungen\_};
\node[black,base right=0cm of node-5-13] (node-5-14) {.\_};
\end{tikzpicture}
}
\end{flushleft}
\caption{An example of longest-first generation.}
\label{fig:width-example}
\end{figure}

\begin{figure}[t]
\small
\begin{CJK*}{UTF8}{gbsn}
\begin{flushleft}
\textbf{Input:} imagine eating enough peanuts to serve as your dinner . \\
\vspace{0.5em}
\textbf{Output:} 想象 一下 ， 吃 足够 的 花生 作为 你 的 晚餐 。 \\
\vspace{0.5em}
\textbf{Parallel decode (common-first):}
\end{flushleft}
\vspace{-0.5em}
\begin{flushleft}
\resizebox{0.75\columnwidth}{!}{%
\begin{tikzpicture}[every node/.style={anchor=base,inner sep=0.1em}]
\node[gray!60] (node-0-0) {想象\_};
\node[gray!60,base right=0cm of node-0-0] (node-0-1) {一下\_};
\node[gray!60,base right=0cm of node-0-1] (node-0-2) { ， \_};
\node[gray!60,base right=0cm of node-0-2] (node-0-3) {吃\_};
\node[gray!60,base right=0cm of node-0-3] (node-0-4) {足够\_};
\node[blue,base right=0cm of node-0-4] (node-0-5) {\uline{的\_}};
\node[gray!60,base right=0cm of node-0-5] (node-0-6) {花};
\node[gray!60,base right=0cm of node-0-6] (node-0-7) {生\_};
\node[gray!60,base right=0cm of node-0-7] (node-0-8) {作为\_};
\node[gray!60,base right=0cm of node-0-8] (node-0-9) {你\_};
\node[gray!60,base right=0cm of node-0-9] (node-0-10) {的\_};
\node[gray!60,base right=0cm of node-0-10] (node-0-11) {晚餐\_};
\node[gray!60,base right=0cm of node-0-11] (node-0-12) { 。\_};
\node[gray!60,below=0.1cm of node-0-0] (node-1-0) {想象\_};
\node[gray!60,base right=0cm of node-1-0] (node-1-1) {一下\_};
\node[blue,base right=0cm of node-1-1] (node-1-2) {\uline{ ， \_}};
\node[gray!60,base right=0cm of node-1-2] (node-1-3) {吃\_};
\node[gray!60,base right=0cm of node-1-3] (node-1-4) {足够\_};
\node[black,base right=0cm of node-1-4] (node-1-5) {的\_};
\node[gray!60,base right=0cm of node-1-5] (node-1-6) {花};
\node[gray!60,base right=0cm of node-1-6] (node-1-7) {生\_};
\node[gray!60,base right=0cm of node-1-7] (node-1-8) {作为\_};
\node[gray!60,base right=0cm of node-1-8] (node-1-9) {你\_};
\node[blue,base right=0cm of node-1-9] (node-1-10) {\uline{的\_}};
\node[gray!60,base right=0cm of node-1-10] (node-1-11) {晚餐\_};
\node[gray!60,base right=0cm of node-1-11] (node-1-12) { 。\_};
\node[gray!60,below=0.1cm of node-1-0] (node-2-0) {想象\_};
\node[blue,base right=0cm of node-2-0] (node-2-1) {\uline{一下\_}};
\node[black,base right=0cm of node-2-1] (node-2-2) { ， \_};
\node[blue,base right=0cm of node-2-2] (node-2-3) {\uline{吃\_}};
\node[gray!60,base right=0cm of node-2-3] (node-2-4) {足够\_};
\node[black,base right=0cm of node-2-4] (node-2-5) {的\_};
\node[gray!60,base right=0cm of node-2-5] (node-2-6) {花};
\node[gray!60,base right=0cm of node-2-6] (node-2-7) {生\_};
\node[gray!60,base right=0cm of node-2-7] (node-2-8) {作为\_};
\node[blue,base right=0cm of node-2-8] (node-2-9) {\uline{你\_}};
\node[black,base right=0cm of node-2-9] (node-2-10) {的\_};
\node[gray!60,base right=0cm of node-2-10] (node-2-11) {晚餐\_};
\node[blue,base right=0cm of node-2-11] (node-2-12) {\uline{ 。\_}};
\node[blue,below=0.1cm of node-2-0] (node-3-0) {\uline{想象\_}};
\node[black,base right=0cm of node-3-0] (node-3-1) {一下\_};
\node[black,base right=0cm of node-3-1] (node-3-2) { ， \_};
\node[black,base right=0cm of node-3-2] (node-3-3) {吃\_};
\node[blue,base right=0cm of node-3-3] (node-3-4) {\uline{足够\_}};
\node[black,base right=0cm of node-3-4] (node-3-5) {的\_};
\node[gray!60,base right=0cm of node-3-5] (node-3-6) {花};
\node[gray!60,base right=0cm of node-3-6] (node-3-7) {生\_};
\node[blue,base right=0cm of node-3-7] (node-3-8) {\uline{作为\_}};
\node[black,base right=0cm of node-3-8] (node-3-9) {你\_};
\node[black,base right=0cm of node-3-9] (node-3-10) {的\_};
\node[blue,base right=0cm of node-3-10] (node-3-11) {\uline{晚餐\_}};
\node[black,base right=0cm of node-3-11] (node-3-12) { 。\_};
\node[black,below=0.1cm of node-3-0] (node-4-0) {想象\_};
\node[black,base right=0cm of node-4-0] (node-4-1) {一下\_};
\node[black,base right=0cm of node-4-1] (node-4-2) { ， \_};
\node[black,base right=0cm of node-4-2] (node-4-3) {吃\_};
\node[black,base right=0cm of node-4-3] (node-4-4) {足够\_};
\node[black,base right=0cm of node-4-4] (node-4-5) {的\_};
\node[gray!60,base right=0cm of node-4-5] (node-4-6) {花};
\node[blue,base right=0cm of node-4-6] (node-4-7) {\uline{生\_}};
\node[black,base right=0cm of node-4-7] (node-4-8) {作为\_};
\node[black,base right=0cm of node-4-8] (node-4-9) {你\_};
\node[black,base right=0cm of node-4-9] (node-4-10) {的\_};
\node[black,base right=0cm of node-4-10] (node-4-11) {晚餐\_};
\node[black,base right=0cm of node-4-11] (node-4-12) { 。\_};
\node[black,below=0.1cm of node-4-0] (node-5-0) {想象\_};
\node[black,base right=0cm of node-5-0] (node-5-1) {一下\_};
\node[black,base right=0cm of node-5-1] (node-5-2) { ， \_};
\node[black,base right=0cm of node-5-2] (node-5-3) {吃\_};
\node[black,base right=0cm of node-5-3] (node-5-4) {足够\_};
\node[black,base right=0cm of node-5-4] (node-5-5) {的\_};
\node[blue,base right=0cm of node-5-5] (node-5-6) {\uline{花}};
\node[black,base right=0cm of node-5-6] (node-5-7) {生\_};
\node[black,base right=0cm of node-5-7] (node-5-8) {作为\_};
\node[black,base right=0cm of node-5-8] (node-5-9) {你\_};
\node[black,base right=0cm of node-5-9] (node-5-10) {的\_};
\node[black,base right=0cm of node-5-10] (node-5-11) {晚餐\_};
\node[black,base right=0cm of node-5-11] (node-5-12) { 。\_};
\end{tikzpicture}
}
\end{flushleft}
\end{CJK*}
\caption{An example of common-first generation.}
\label{fig:frequency-example}
\end{figure}

\section{Experiments}
\label{sec:experiments}

\begin{table*}[t]
\centering
\begin{tabular}{lccccccc}
\toprule
\multirow[t]{2}{*}{\bfseries Order} & & \multicolumn{3}{c}{\bfseries En $\to$ De} & \multicolumn{3}{c}{\bfseries En $\to$ Zh} \\
\cmidrule{2-8}
& $\tau$ & $0.5$ & $1.0$ & $2.0$ & $0.5$ & $1.0$ & $2.0$ \\
\midrule
Binary Tree & & 91\% & 86\% & 80\% & 88\% & 83\% & 78\% \\
Random & & 86\% & 81\% & 72\% & 82\% & 77\% & 68\% \\
Left-to-Right & & 95\% & 88\% & 77\% & 88\% & 82\% & 70\% \\
Right-to-Left & & 95\% & 90\% & 78\% & 92\% & 83\% & 72\% \\
Common First & & 92\% & 88\% & 80\% & 88\% & 84\% & 76\% \\
Rare First & & 88\% & 81\% & 73\% & 83\% & 77\% & 67\% \\
Shortest First & & 93\% & 88\% & 80\% & 91\% & 84\% & 76\% \\
Longest First & & 92\% & 86\% & 77\% & 92\% & 84\% & 76\% \\
Alphabetical (A $\to$ z) & & 93\% & 87\% & 77\% & 88\% & 82\% & 73\% \\
Alphabetical (z $\to$ A) & & 90\% & 84\% & 74\% & 85\% & 78\% & 69\% \\
\bottomrule
\end{tabular}
\caption{Percentage of insertions that follow the target order exactly, averaged over the development set.}
\label{table:insertion-accuracy}
\end{table*}

For our experiments, we train and evaluate models for each order on two standard machine translation datasets: WMT14 En-De and WMT18 En-Zh. For WMT14 En-De, we follow the standard setup with newstest2013 as our development set and newstest2014 as our test set. For WMT18 En-Zh, we use the official preprocessed data\footnote{http://data.statmt.org/wmt18/translation-task/preprocessed/zh-en/} with no additional data normalization or filtering, taking newstest2017 to be our development set and newstest2018 our test set. En-Zh evaluation is carried out using sacreBLEU\footnote{BLEU+case.mixed+lang.en-zh+numrefs.1+smooth.exp+test.wmt18+tok.zh+version.1.2.12} \cite{post-wmt-2018}. In both cases, we train all models for 1M steps using sequence-level knowledge distillation \cite{hinton-nips-2015,kim-emnlp-2016} from a base Transformer \cite{vaswani-nips-2017}. We perform a sweep over temperatures $\tau \in \{0.5, 1, 2\}$ and EOS penalties $\in \{0, 0.5, 1, 1.5, \dots, 8\}$ \cite{stern-arxiv-2019} on the development set, but otherwise perform no additional hyperparameter tuning, borrowing all other model and optimization settings from the base Transformer.

\subsection{Ability to Learn Different Orders}

By and large, we find that the Insertion Transformer is remarkably capable of learning to generate according to whichever order it was trained for. We give example decodes for three different generation orders in Figures \ref{fig:alphabetical-examples}, \ref{fig:width-example}, and \ref{fig:frequency-example}. In the first example, we see that the alphabetical En-De model adheres to the Unicode ordering for Latin characters (punctuation $\to$ uppercase $\to$ lowercase), and that the En-Zh model similarly adheres to the Unicode order for Chinese (punctuation $\to$ CJK characters sorted by radical and stroke count).
In the second example, the longest-first En-De model generates subwords in decreasing order of length as expected. Finally, in the third example, the common-first En-Zh model begins with common particles and punctuation before generating the main content words.

We give a quantitative measurement of the success of each model in Table \ref{table:insertion-accuracy}, computing the percentage of insertions across the development set that adhered to the best-scoring action under the desired ordering. Most models exhibit similar trends, with the majority of En-De models achieving accuracies in excess of 90\% when a low temperature is used, and with corresponding results in the mid-to-upper 80\% range for En-Zh. Even the random order based on token hashes has accuracies exceeding 80\% for both languages, demonstrating that the model has a strong capacity to adapt to any oracle policy.

\begin{table*}[t]
\centering
\begin{tabular}{lcccc}
\toprule
\multirow[t]{3}{*}{\bfseries Order} & \multicolumn{2}{c}{\bfseries En $\to$ De} & \multicolumn{2}{c}{\bfseries En $\to$ Zh} \\
\cmidrule{2-5}
& \bfseries Serial & \bfseries Parallel & \bfseries Serial & \bfseries Parallel \\
\midrule
& \multicolumn{2}{c}{\citet{vaswani-nips-2017}} & \multicolumn{2}{c}{This Work} \\
Transformer & 27.3 & & 35.8 & \\
\midrule
& \multicolumn{2}{c}{\citet{stern-arxiv-2019}} & \multicolumn{2}{c}{This Work} \\
Uniform & 27.12 & 26.72 & 32.9 & 33.1 \\
Binary Tree & 27.29 & 27.41 & 32.6 & 34.0 \\
\midrule
& \multicolumn{4}{c}{This Work} \\
Random & 26.15 & 26.10 & 32.6 & 32.4 \\
Left-to-Right & 26.37 & 25.56 & 31.7 & 31.2 \\
Right-to-Left & 26.60 & 24.49 & 32.4 & 30.8 \\
Common First & 26.88 & 26.86 & 33.5 & 32.9 \\
Rare First & 26.06 & 26.24 & 32.5 & 32.2 \\
Shortest First & 27.05 & 27.15 & 33.0 & 32.7 \\
Longest First & 26.45 & 26.41 & 32.8 & 33.2 \\
Alphabetical (A $\rightarrow$ z) & 26.86 & 26.58 & 32.7 & 32.5 \\
Alphabetical (z $\rightarrow$ A) & 27.22 & 26.37 & 33.1 & 33.0 \\
Easy First & 26.95 & 27.05 & 32.5 & 32.5 \\
Hard First & 25.85 & 26.30 & 32.4 & 32.9 \\
\bottomrule
\end{tabular}
\caption{Test BLEU results for WMT14 En-De newstest2014 and WMT18 En-Zh newstest2018 with serial and parallel decoding.}
\label{table:test-results}
\end{table*}

\begin{figure*}[t]
\centering
\mbox{
\subfigure[English $\to$ German]{\includegraphics[width=0.5\textwidth]{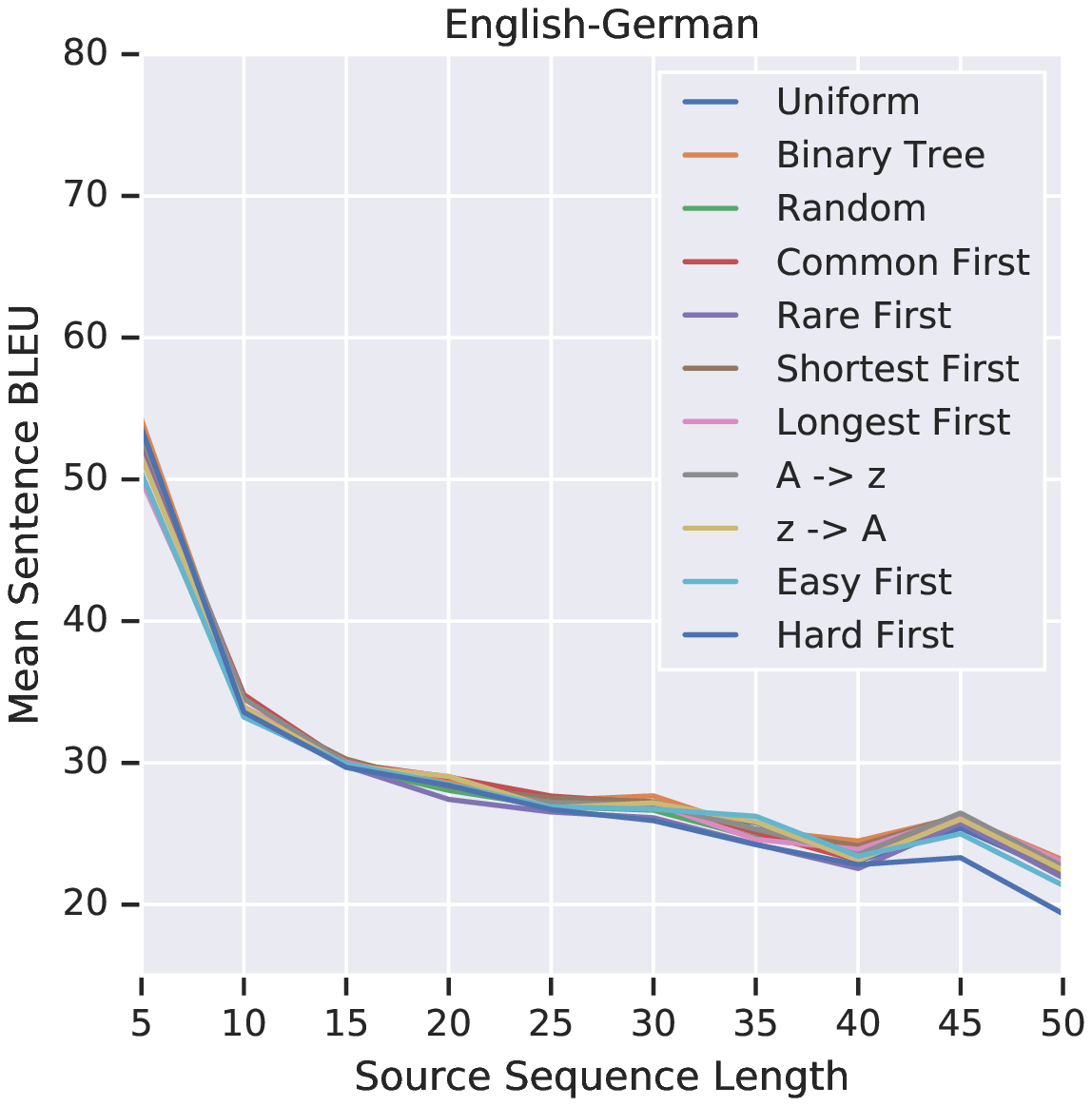}}
\subfigure[English $\to$ Chinese]{\includegraphics[width=0.5\textwidth]{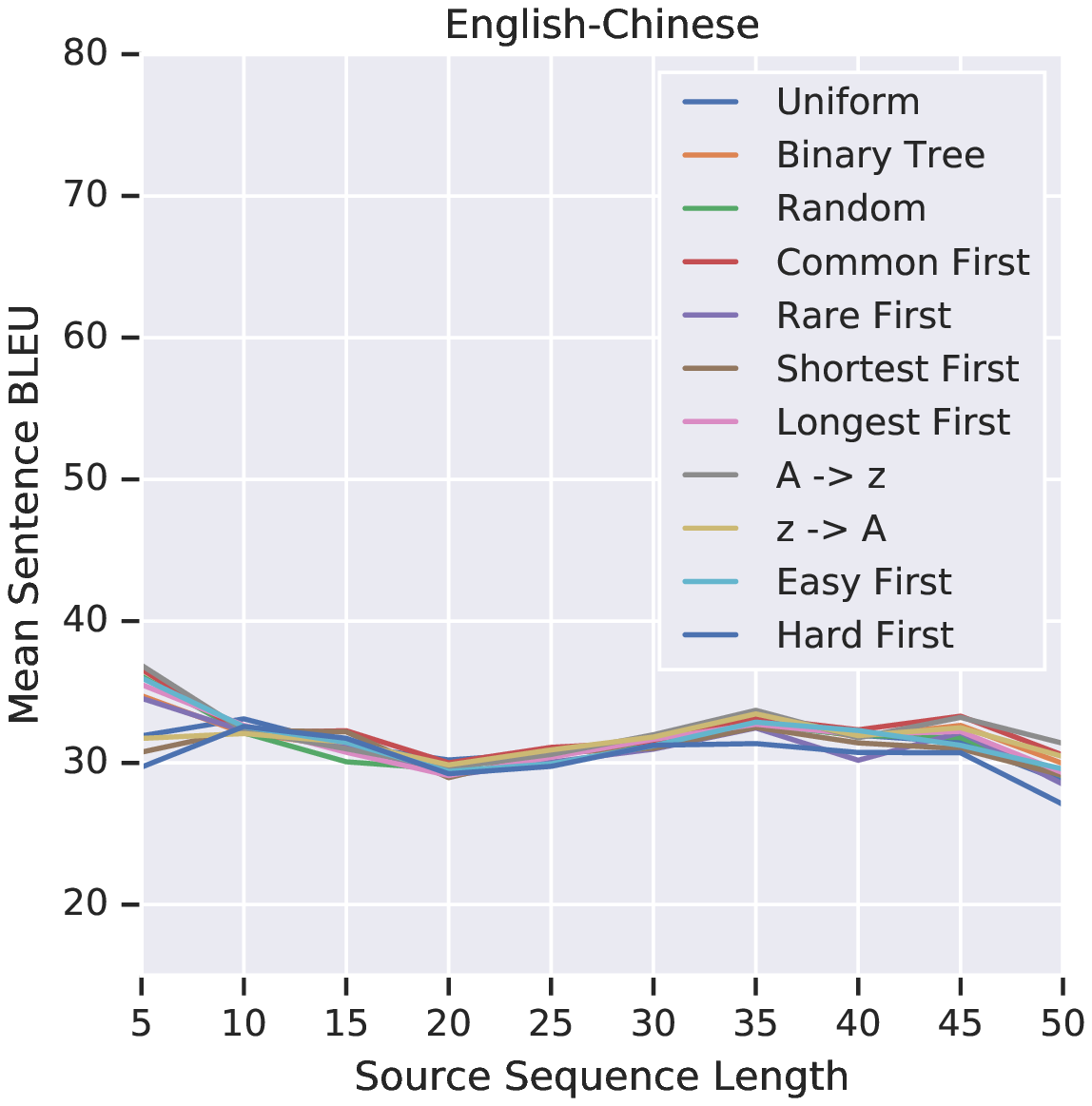}}
}
\caption{Sentence-level BLEU scores as a function of sentence length for several of our model variants. Source sentences in each development set are binned into groups of size 5, up to length 50.}
\label{fig:bleubin}
\end{figure*}

\subsection{Test Results}

Next, we measure the quality of our models by evaluating their performance on their respective test sets. The BLEU scores are reported in Table \ref{table:test-results}. The uniform loss proposed by \citet{stern-arxiv-2019} serves as a strong baseline for both language pairs, coming within 0.6 points of the original Transformer for En-De at 26.72 BLEU, and attaining a respectable score of 33.1 BLEU on En-Zh. We note that there is a slightly larger gap between the normal Transformer and the Insertion Transformer for the latter of 2.7 points, which we hypothesize is a result of the larger discrepancy between word orders in the two languages combined with the more difficult nature of the Insertion Transformer training objective.

Most of the content-based orderings (frequency-based, length-based, alphabetical) perform comparably to the uniform loss, and even the random order is not far behind. The adaptive orders perform similarly well, with easy-first attaining one of the highest scores on En-De. Curiously, in our model adaptive easy-order, we were unable to identify any strong patterns in the generation order. The model did have a slight preference towards functional words (i.e., ``,'' and ``der''), but the preference was weak. As for location-based losses, the binary tree loss is notable in that it achieves the highest score across all losses for both languages. On the other hand, we note that while the soft left-to-right and right-to-left losses perform substantially better than the hard loss employed in the original work by \citet{stern-arxiv-2019}, performance does suffer when using parallel decoding for those models, which is generally untrue of the other orderings. We believe this is due in part to exposure bias issues arising from the monotonic ordering as compared with the uniform roll-in policy that are not shared by the other losses.

\subsection{Performance vs.\ Sentence Length}

For additional analysis, we consider how well our models perform relative to one another conditional on the length of the source sentence. Sentence length can be seen as a rough proxy measurement of the difficulty of translating a sentence. This is to determine if whether some order variations are able to achieve improved BLEU scores over other models depending on the source sentence's length. For each sentence in the En-De and En-Zh development sets, we compute their lengths and bin them into groups of size 5, up to a maximum length of 50. Within each bin, we compute sentence-level BLEU and take the mean score across all sentences. This is done for each of our model variants. Figure~\ref{fig:bleubin} illustrates the results of this experiment. We observe a surprisingly small model variance across all bin lengths. This suggests that sentences that are difficult to translate are difficult across all orderings, and no particular ordering appears strictly better or worse than others. One small exception to this is a performance fall-off of hard-first orderings for very long sentences across both datasets. We also observe a different distribution of BLEU scores across bin lengths for En-De and En-Zh. In particular, En-De models are approximately monotonic-decreasing in performance as source length increases, while on En-Zh performance is roughly flat across sentence length. This also highlights the importance of taking additional diverse language pairs into consideration, as translation properties on one language pair may not be observed in others.

Ultimately, given the similarity of the development scores across sentence lengths and the test scores for the various models, we come to the surprising conclusion that for single-sentence English-German translation, generation order is relatively unimportant. However, for English-Chinese, it is unclear and we leave further analysis to future work. Under the Insertion Transformer framework, it appears order also does not matter much, however there is a 2.7 BLEU gap between the results in the Insertion Transformer and our Transformer baseline.

\section{Related Work}

In recent work, several insertion-based frameworks have been proposed for the generation of sequences in a non-left-to-right fashion for machine translation \cite{stern-arxiv-2019,welleck-arxiv-2019,gu-arxiv-2019}. \citet{stern-arxiv-2019} introduced the Insertion Transformer and explored uniform and balanced binary tree orders. We built upon and generalized this approach in order to explore a much broader set of orders. \citet{welleck-arxiv-2019} explored insertions using a binary-tree formulation. They also explored uniform and model-based orders, but found them to lag significantly behind their left-to-right baselines. Additionally, despite using a binary-tree formulation for generation, they did not explore tree-based orders. \citet{gu-arxiv-2019} introduced a model which did not explicitly represent the output canvas arising from insertions, but rather used an implicit representation through conditioning on the insertion sequence. They also performed an exploration of different generation orders, including random, odd-even, common-first, rare-first, and a search-adaptive order. Their search-adaptive order can be seen as a global version of our local model adaptive order, where we use the local greedy posterior as the reward function, and they use the sequence level log-probability as the reward function. Curiously, in their framework, the random order fell significantly behind the left-to-right baseline, while they showed small gains in their search adaptive order. One key difference between our work and \citet{welleck-arxiv-2019} and \citet{gu-arxiv-2019} is that we use a soft order-reward framework as opposed to teacher forcing. This might explain some of the performance differences, as our framework allows for a more flexible training objective. Additionally, since we use a uniform roll-in policy, our models may have less of a label bias problem, as they are trained to be able to continue from any partial output rather than just those arising from the target policy.

\section{Conclusion}

In this work, we investigated a broad array of generation orders for machine translation using an insertion-based sequence generation model, the Insertion Transformer. We found that regardless of the type of strategy selected, be it location-based, frequency-based, length-based, alphabetical, model-based, or even random, the Insertion Transformer is able to learn it with high fidelity and produce high-quality output in the selected order. This is especially true for English-German single sentence translation, in which we by and large found order to not matter. This opens a wide range of possibilities for generation tasks where monotonic orderings are not the most natural choice, and we would be excited to explore some of these areas in future work.

\bibliography{acl2019}
\bibliographystyle{acl_natbib}

\clearpage
\onecolumn
\section*{Appendix}

Full development set results for En-De translation and En-Zh translation.

\begin{table*}[h]
\centering
\small
\begin{tabular}{lccccc}
\toprule
 \multirow[t]{3}{*}{\bfseries Order} & & \multicolumn{2}{c}{\bfseries English $\rightarrow$ German} & \multicolumn{2}{c}{\bfseries English $\rightarrow$ Chinese} \\
 \cmidrule{2-6}
 & $\tau$ & \bfseries BLEU (+EOS) & \bfseries BLEU (+EOS) & \bfseries BLEU (+EOS) & \bfseries BLEU (+EOS) \\
\cmidrule{2-6}
& & & \bfseries +Parallel & & \bfseries +Parallel \\
\midrule
& & \multicolumn{2}{c}{\citet{stern-arxiv-2019}} & \multicolumn{2}{c}{This Work} \\
Uniform & $\infty$ & 22.39 (25.58) & 24.31 (24.91) & 28.6 (31.8) & 30.4 (31.9) \\
\midrule
\multirow[t]{3}{*}{Binary Tree}
& $0.5$ & 24.49 (25.55) & 25.33 (25.70) & 29.3 (31.6) & 31.3 (31.9) \\
& $1.0$ & 24.36 (25.43) & 25.43 (25.76) & 29.6 (32.0) & 31.4 (32.2) \\
& $2.0$ & 24.59 (25.80) & 25.33 (25.80) & 29.1 (32.2) & 31.4 (32.3) \\
\midrule
& & \multicolumn{4}{c}{This Work} \\
\multirow[t]{3}{*}{Random}
& $0.5$ & 23.82 (24.87) & 23.97 (24.20) & 28.5 (30.6) & 29.4 (30.2) \\
& $1.0$ & 24.03 (25.46) & 24.58 (24.82) & 28.6 (31.1) & 30.0 (31.0) \\
& $2.0$ & 24.00 (25.41) & 24.68 (25.07) & 28.9 (31.7) & 30.4 (31.6)  \\
\midrule
\multirow[t]{3}{*}{L2R (Left-Aligned)}
& $0.5$ & 21.19 (24.46) & 21.40 (21.57) & 24.5 (30.0) & 25.7 (28.3) \\
& $1.0$ & 21.36 (24.02) & 20.84 (21.25) & 24.8 (29.8) & 25.2 (27.8) \\
& $2.0$ & 21.78 (24.21) & 20.56 (21.11) & 25.8 (29.8) & 24.9 (27.6) \\
\multirow[t]{3}{*}{L2R (Right-Aligned)}
& $0.5$ & 21.77 (25.00) & 22.62 (23.38) & 25.6 (31.6) & 27.3 (30.0) \\
& $1.0$ & 21.85 (25.22) & 22.78 (23.67) & 25.3 (31.2) & 27.0 (30.1) \\
& $2.0$ & 21.01 (24.88) & 22.29 (23.80) & 23.5 (30.9) & 25.8 (30.4) \\
\multirow[t]{3}{*}{R2L (Left-Aligned)}
& $0.5$ & 23.75 (25.04) & 23.15 (23.25) & 27.6 (31.4) & 27.8 (28.6) \\
& $1.0$ & 23.72 (25.29) & 22.89 (22.89) & 28.0 (31.6) & 28.0 (29.3) \\
& $2.0$ & 24.09 (25.64) & 23.61 (23.85) & 28.6 (31.9) & 28.3 (29.9) \\
\multirow[t]{3}{*}{R2L (Right-Aligned)}
& $0.5$ & 19.23 (23.52) & 19.70 (21.02) & 21.3 (31.3) & 22.3 (28.3) \\
& $1.0$ & 19.56 (23.27) & 20.20 (21.55) & 20.9 (30.5) & 21.6 (28.3) \\
& $2.0$ & 20.19 (23.55) & 20.84 (22.22) & 20.3 (30.9) & 21.5 (28.7) \\
\midrule
\multirow[t]{3}{*}{Common First}
& $0.5$ & 25.20 (25.43) & 25.05 (25.05) & 29.9 (31.2) & 30.5 (30.5) \\
& $1.0$ & 25.46 (25.84) & 25.76 (25.81) & 30.5 (32.0) & 31.1 (31.3) \\
& $2.0$ & 25.30 (25.76) & 25.75 (25.83) & 30.4 (32.2) & 31.4 (31.9) \\
\multirow[t]{3}{*}{Rare First}
& $0.5$ & 22.83 (24.30) & 23.19 (23.62) & 27.0 (29.5) & 28.7 (29.7) \\
& $1.0$ & 22.75 (24.56) & 23.42 (23.99) & 27.9 (30.7) & 29.5 (30.5) \\
& $2.0$ & 23.10 (24.79) & 24.00 (24.36) & 28.1 (31.2) & 29.8 (31.1) \\
\midrule
\multirow[t]{3}{*}{Shortest First}
& $0.5$ & 24.93 (25.55) & 24.94 (25.01) & 27.4 (30.3) & 29.1 (30.0) \\
& $1.0$ & 24.95 (25.72) & 25.17 (25.28) & 28.0 (30.9) & 29.6 (30.8) \\
& $2.0$ & 25.05 (25.85) & 25.26 (25.48) & 28.2 (31.4) & 30.3 (31.5) \\
\multirow[t]{3}{*}{Longest First}
& $0.5$ & 23.59 (25.09) & 24.24 (24.56) & 29.2 (31.4) & 30.5 (31.2) \\
& $1.0$ & 23.53 (25.07) & 24.68 (25.13) & 29.2 (31.5) & 31.0 (31.8) \\
& $2.0$ & 24.09 (25.78) & 24.93 (25.37) & 29.0 (31.9) & 31.1 (32.1) \\
\midrule
\multirow[t]{3}{*}{Alphabetical (A $\rightarrow$ Z $\rightarrow$ a $\rightarrow$ z)}
& $0.5$ & 24.49 (25.15) & 24.87 (24.91) & 29.2 (31.0) & 30.1 (30.6) \\
& $1.0$ & 24.61 (25.19) & 24.96 (25.12) & 30.1 (32.0) & 30.8 (31.4) \\
& $2.0$ & 24.77 (25.67) & 25.45 (25.71) & 29.7 (32.1) & 30.7 (31.8) \\
\multirow[t]{3}{*}{Alphabetical (z $\rightarrow$ a $\rightarrow$ Z $\rightarrow$ A)}
& $0.5$ & 24.16 (25.24) & 24.56 (24.73) & 29.2 (31.4) & 30.3 (30.8) \\
& $1.0$ & 24.19 (25.45) & 24.65 (25.10) & 29.3 (31.9) & 30.7 (31.5) \\
& $2.0$ & 24.26 (25.76) & 25.02 (25.40) & 29.7 (32.3) & 31.0 (32.0) \\
\midrule
\multirow[t]{3}{*}{Easy First}
& $0.5$ & 22.58 (24.09) & 22.16 (22.63) & 27.5 (30.2) & 28.4 (29.7) \\
& $1.0$ & 23.68 (25.08) & 23.66 (24.03) & 28.9 (31.6) & 29.3 (30.7) \\
& $2.0$ & 23.87 (25.43) & 24.64 (25.26) & 29.1 (31.9) & 30.4 (31.7) \\
\multirow[t]{3}{*}{Hard First}
& $0.5$ & 20.01 (23.46) & 23.16 (23.61) & 24.7 (29.7) & 28.7 (30.2) \\
& $1.0$ & 20.96 (24.36) & 23.76 (24.56) & 25.4 (30.1) & 29.1 (30.7) \\
& $2.0$ & 21.97 (24.90) & 24.33 (24.70) & 26.4 (31.1) & 29.9 (31.4) \\
\bottomrule
\end{tabular}
\caption{Development BLEU results for WMT14 En-De newstest2013 and WMT18 En-Zh newstest2017. The first number in each column is the result obtained without an EOS penalty, while the second number in parentheses is the score obtained with the best EOS penalty for that setting.}
\vspace{-6em}
\end{table*}

\end{document}